\documentclass{article}
\usepackage{ijcai24}

\usepackage{times}
\usepackage{soul}
\usepackage{url}
\usepackage[hidelinks]{hyperref}
\usepackage[utf8]{inputenc}
\usepackage[small]{caption}
\usepackage{graphicx}
\usepackage{amsmath}
\usepackage{amsthm}
\usepackage{booktabs}
\usepackage{algorithm}
\usepackage{algorithmic}
\usepackage[switch]{lineno}
\usepackage{cleveref}
\usepackage{fancyhdr}

\title{Addressing Image Hallucination in Text-to-Image Generation through\\ Factual Image Retrieval}

\author{
Youngsun Lim
\and
Hyunjung Shim\\
\affiliations
Kim Jaechul Graduate School of AI, KAIST\\
\emails
\{youngsun\_ai, kateshim\}@kaist.ac.kr,
}

\begin{document}

\maketitle

\begin{figure*}[t!]
\centering
\includegraphics[width=1.0\textwidth]{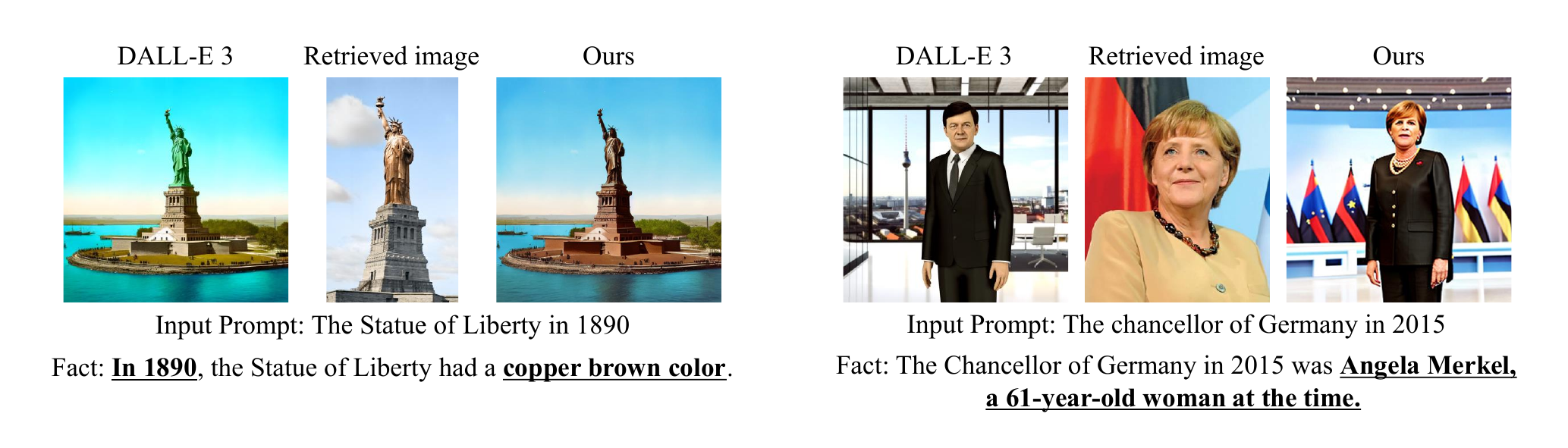}
\vspace{-0.3in}
\caption{Examples of image hallucination and the facts that should have been reflected.}
\vspace{-1.0em}
\label{fig1} 
\end{figure*}


\begin{abstract}
    Text-to-image generation has shown remarkable progress with the emergence of diffusion models. However, these models often generate factually inconsistent images, failing to accurately reflect the factual information and common sense conveyed by the input text prompts. We refer to this issue as \emph{Image hallucination}. Drawing from studies on hallucinations in language models, we classify this problem into three types and propose a methodology that uses factual images retrieved from external sources to generate realistic images. Depending on the nature of the hallucination, we employ off-the-shelf image editing tools, either InstructPix2Pix or IP-Adapter, to leverage factual information from the retrieved image. This approach enables the generation of images that accurately reflect the facts and common sense.
\end{abstract}

\section{Introduction}

Recent text-to-image generation methods have made remarkable progress due to the emergence of diffusion models. However, many text-to-image generation models still fail to accurately reflect the underlying facts or changes over time and circumstances conveyed by input text prompts. As a result, they generate prototypical examples that differ from reality. For instance, the Statue of Liberty, completed in 1886, initially had a copper-brown color because its surface was covered with copper. Over the decades, the color gradually changed to its current blue-green hue due to oxidation. However, when the prompt ``The Statue of Liberty in 1890'' is entered into a recent text-to-image model such as Dall-E 3 \cite{betker2023improving}, only the turquoise Statue of Liberty is generated, as shown in Figure~\ref{fig1}.

Generating inaccurate images can spread misinformation and misconceptions \cite{verge24}. It poses a serious issue in fields where factual accuracy is crucial, such as education and journalism. 
Recent practices of utilizing foundation models, such as Stable Diffusion \cite{StableDiffusion} or Imagen \cite{imagen}, have become prevalent in text-to-image generation. However, inaccurate outputs from these models could propagate biases and distortions into subsequent models. 
Additionally, the reliability of AI models hinges on their ability to deliver accurate results. While large language models (LLMs) have become widely used in industry, text-to-image generation models still lack practical usability. This may be because generated images do not yet accurately reflect factual information. Therefore, it is essential for image generation models to produce factually accurate and trustworthy images when applied to the aforementioned applications.

Despite its significance, the prevention of generating inaccurate images has been underestimated, and there is no appropriate term to refer to this issue in the context of text-to-image generation. In this study, we define this problem as “image hallucination”.
Image hallucination includes not only misalignment between text prompts and generated images but also the generation of factually defective images. This concept is more complex than alignment because it involves understanding meanings and facts not included in the text prompt itself. In this study, we focus on the issue of text-to-image generation failing to generate facts.

Because image hallucination is extensive, it is difficult to address all issues. Thus, we highlight the hallucinations that are judged to be representative based on \cite{survey} by categorizing them into three types: (1) Factual inconsistency caused by co-occurrence bias, (2) outdated knowledge hallucination, and (3) factual fabrication that produces counterfactual. We will discuss in detail the three types of hallucinations that will be addressed later.

To solve the previously mentioned problem, external knowledge can be utilized to guide the image generation model.
Recently, retrieval-augmented language models have demonstrated significant potential in addressing hallucination \cite{NLPimprove,NLPimprove2}. Given input text, such models retrieve relevant documents from external memory and generate fact-based answers. Recent study \cite{racm3} has expanded retrieval and generation to encompass both images and text, training multimodal generation models to effectively utilize retrieved information.

We develop this idea and introduce a tuning-free method to enhance the image generation model for producing fact-based images.
Initially, an image is generated via an existing text-to-image model. Then, the input prompt is used as the search query to retrieve $N$ images in order of relevance. Among the retrieved images, the user selects one factual image to be used as guidance to eliminate image hallucination. Depending on the type of hallucination, we propose two methods. (1) If the hallucination affects a certain property, InstructPix2Pix \cite{instructpix2pix} is used to address it. Referring to the procedure of generating instructions from \cite{instructpix2pix}, the generated and retrieved images are input into an LLM (e.g. GPT-4) to generate instructions based on differences between the two images. These instructions, along with the initially generated image, are entered into the pre-trained InstructPix2Pix to produce an image with the hallucination corrected. (2) If a hallucination occurs over a wide and complex area, IP-Adapter \cite{ipadapter} is utilized to remove it. The input text prompt and the retrieved image are fed into the LLM to generate text prompts depicting the retrieved image. These generated prompts, along with the retrieved image and the initial generated image, are input into the pre-trained IP-Adapter to eliminate the hallucination in the initial generated image.

Through this approach, we can eliminate hallucinations in existing text-to-image generation models without incurring training costs. Additionally, users can reflect their intentions and enhance the trustworthiness of the edited image by interactively selecting the retrieved image from the search results.

\begin{figure*}[t!]
\centering
\includegraphics[width=0.9\textwidth]{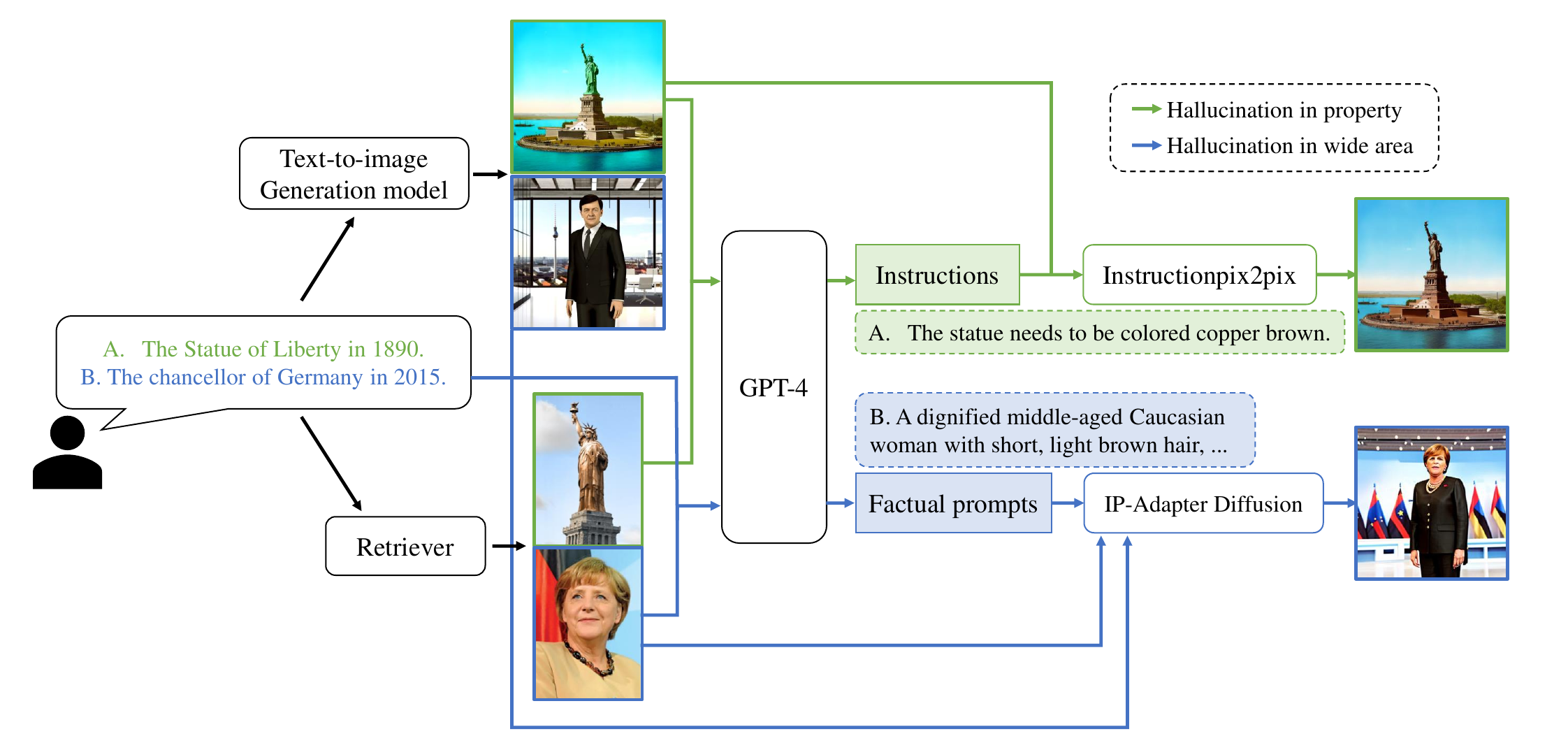}
\vspace{-0.1in}
\caption{The overall pipeline indicates two different strategies for preventing image hallucination based on the target of the hallucination.}
\label{fig:fig2} 
\end{figure*}

\section{Related Work}
Text-to-image diffusion models \cite{StableDiffusion,imagen} have made significant progress but often struggle with complex prompts. The early approach utilized additional inputs, such as key points, to achieve better control \cite{yang2023paint}. Recent advancements leverage LLMs to manage layout directly \cite{wu2023self}, thereby improving prompt alignment. Diffusion models enable various image edits, ranging from global styles to precise object manipulation, but often lack precision for detailed spatial adjustments \cite{hertz2022prompt}. We address this issue by utilizing images from external sources to create fact-based images.

Some research focuses on generative models trained to retrieve information in multimodal settings. For example, Re-Imagen \cite{reimagen} generates images from retrieved images with text prompts, and MuRAG \cite{murag} generates language answers using retrieved images. Unlike these, our approach does not need any training.

Various strategies exist to mitigate hallucination in LLMs. Data-related issues can be addressed by enhancing data quality and employing better labeling techniques \cite{lin2021truthfulqa}. For training-related hallucinations, improved model architectures and advanced regularization techniques \cite{regular} are recommended. In this paper, we use fact-based images to eliminate hallucinations.

\section{Methods}

\subsection{Image Hallucination}

We focus on the problem of hallucination where the generated image does not reflect the common sense and facts conveyed by the text, rather than a simple misalignment between the text prompt and the generated image. Based on the paper analyzing hallucinations in language models \cite{survey}, we categorize representative image hallucinations into three categories: factual inconsistency caused by co-occurrence bias, Outdated knowledge hallucination that fails to reflect up-to-date information, and factual fabrication where the generated image has little to no basis in reality.

\emph{Factual inconsistency} refers to situations where the output of a generation model contains facts based on real-world information but is contradictory or inaccurate. Specifically, factual inconsistency caused by co-occurrence bias occurs because generation models rely on patterns in the training data, which may be imbalanced. For instance, although the Statue of Liberty was originally copper brown, models generate it with its current bluish-green hue due to the predominance of such data, ignoring its historical appearance.

\emph{Outdated knowledge hallucination} occurs due to the inability to reflect temporal information.
The internal knowledge of a generation model is not updated once trained. Therefore, external information must be used to generate new, updated knowledge over time. For example, current text-to-image generation models cannot accurately generate images of presidents from specific historical periods.

\emph{Factual fabrication} generates scenarios unlikely or impossible when compared to reality. For example, San Francisco rarely experiences snow in the winter, having only witnessed it three times since the 20th century. However, if the prompt is ``The Golden Gate Bridge in winter'', an image with a significant amount of snow is generated.

\begin{figure*}[ht]
\centering
\includegraphics[width=1.0\textwidth]{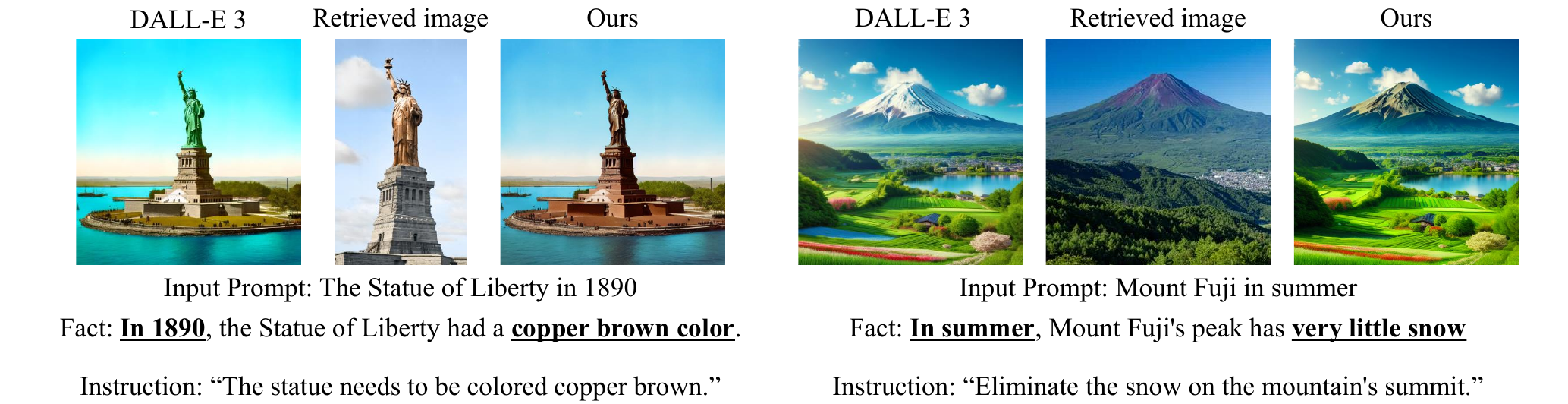}
\vspace{-0.5em}
\caption{Examples showing image hallucination due to factual inconsistency caused by co-occurrence bias, and images resolved by applying our methodology. Instruction is created and utilized using the input prompt and retrieved factual image.}
\label{fig:fig3} 
\vspace{-0.7em}
\end{figure*}

\begin{figure*}[ht]
\centering
\includegraphics[width=1.0\textwidth]{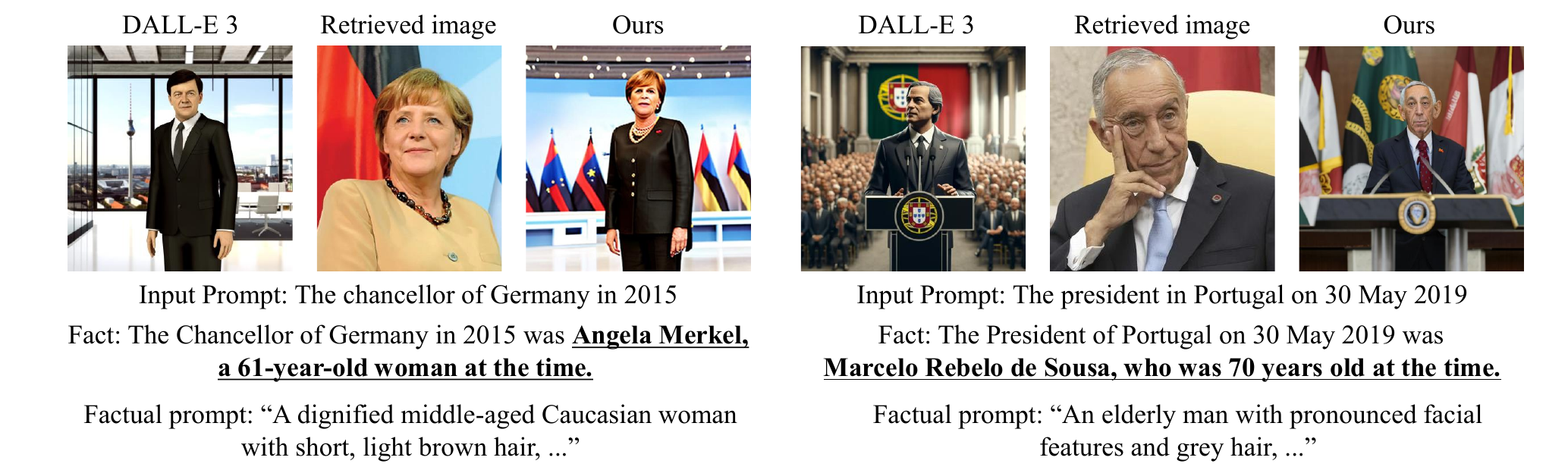}
\caption{Examples showing outdated knowledge hallucination caused by failure to reflect time-shift information and images resolved by applying our methodology. A factual prompt is generated and utilized using the input prompt and the retrieved factual image.}
\label{fig:fig4} 
\vspace{-1.0em}
\end{figure*}

\begin{figure}[ht]
\centering
\includegraphics[width=0.5\textwidth]{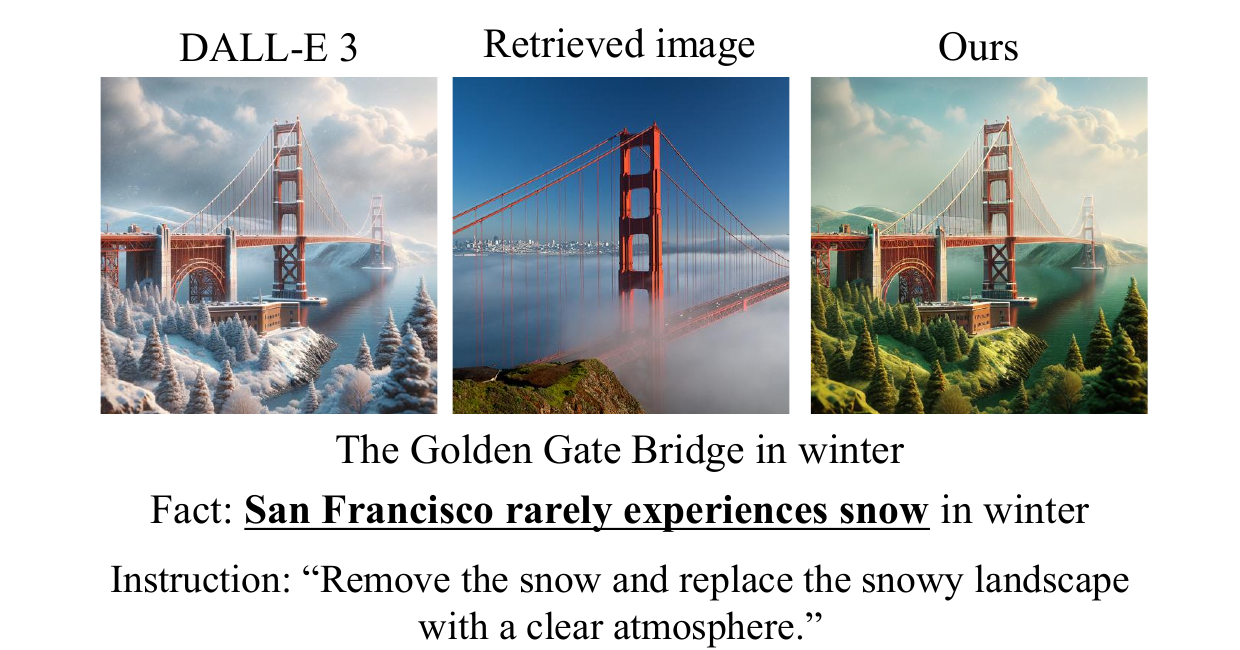}
\caption{Examples showing image hallucination due to factual fabrication, and images resolved by applying our methodology. We applied the same methodology as shown in Figure 3.}
\label{fig:fig5} 
\vspace{-1.0em}
\end{figure}

\subsection{Retrieval-augmented Factual Text-to-Image Generation}
We propose two pipelines, as shown in Figure \ref{fig:fig2}. They utilize retrieved images to reflect real-world knowledge and common sense that generative models cannot capture based on text prompts alone.

\subsubsection{Image Retrieval Interaction}
To search for factual information relevant to a given text prompt, we employ Google's Custom Search JSON API to retrieve images. Among the retrieved images, the user selects the one that best represents the factual information they wish to generate. The number of images retrieved per prompt is a hyperparameter that can be adjusted based on individual requirements. If the desired image is not retrieved for a given prompt, increasing this number facilitates a broader selection of candidate images.

\subsubsection{Overall pipeline}
We propose two methodologies that utilize the retrieved images, depending on the target of the hallucination. First, if a hallucination occurs in a specific property (e.g., the color of an object), we use the retrieved image to obtain instructions and apply InstructPix2Pix. InstructPix2Pix, which combines the knowledge of LLM (GPT-3) and text-to-image generation model (Stable Diffusion), edits images according to human instructions. To train the model, it needs instruction and paired images from paired captions(input and edited caption). The instruction and edited caption are obtained by inputting the initial caption into LLM (GPT-3). Then, the paired captions are entered into the diffusion model to get corresponding images. Based on this, we input the initially generated image and the retrieved image into GPT-4 to generate an instruction based on differences between the two images. For example, the most significant difference between the retrieved image of ``The Statue of Liberty in 1890'' and the generated image is the color. Thus, we input both images into the LLM and obtain the instruction, ``The statue needs to be colored copper brown.'' We input this instruction and the initial image into the pre-trained InstructPix2Pix to correct the hallucination.

On the other hand, if hallucination occurs in a broad area involving many complex subjects, such as a person (including components like the face, hair, clothing, etc.), text prompts or text instructions alone may not be sufficient as generation conditions. Therefore, to incorporate features that cannot be fully expressed in text, the retrieved image must be used as a prompt. For this purpose, we utilize the IP-Adapter, which can process both image and text prompts in combination with a pre-trained diffusion model. The IP-Adapter uses a decoupled cross-attention structure to process image and text features separately. Through this, elaborate images can be created using not only the text prompt but also the image prompt. To efficiently utilize these prompts, we input the text prompt and the retrieved image into LLM (GPT-4) to create a factual prompt depicting features in the retrieved image. This newly generated prompt and the retrieved image are input into the IP-Adapter along with the initially generated image to perform image-to-image editing. This process ultimately removes the hallucination from the initially generated image and edits it to reflect the features of the retrieved image.

\section{Experiments and Results}

We utilize DALL-E 3 as the model for initial image generation. GPT-4 is employed as the LLM that generates instructions and prompts. The InstructPix2Pix and IP-Adapter models are pre-trained models based on Stable Diffusion v1.5.

Figure~\ref{fig:fig3} illustrates hallucinations caused by factual inconsistencies due to co-occurrence bias, and presents the corresponding experimental results. In each example, the input prompt and the retrieved image (middle image) are compared with DALL-E 3's initial output (left image). The initial generation fails to accurately reflect the facts. By providing instructions derived from the discrepancies between the initial and retrieved images to InstructPix2Pix, factually accurate images are obtained (right image). 
For instance, the Statue of Liberty is correctly depicted in its copper brown color as it appeared in 1890, and Mt. Fuji in summer is realistically shown with most of the snow melted from its peak.

Figure~\ref{fig:fig4} demonstrates examples of outdated knowledge hallucinations and the experimental results. When the target of such hallucinations possesses complex and diverse factual information, such as a person, rectifying these inaccuracies through text alone is particularly challenging. Therefore, factual images from search results (middle image) are used directly as prompts. By inputting the retrieved image and a corresponding factual prompt into the IP-Adapter, images that accurately reflect factual information about individuals are generated (right image). Using our methodology, we successfully produce images that accurately depict Angela Merkel as the female Chancellor of Germany in 2015, and Marcelo Rebelo de Sousa as the President of Portugal in May 2019.

Figure~\ref{fig:fig5} depicts hallucinations and experimental results concerning factual fabrication. The initial image does not properly reflect the fact that San Francisco rarely experiences snow in winter. To address this, the same method used for factual inconsistency caused by co-occurrence bias is applied. Instructions obtained from GPT-4, directing the removal of all snow from the initial generation, along with the initial image, are input into InstructPix2Pix. This process generates an image reflecting the factual information of San Francisco with no snow.

\section{Future works}
We plan to expand our research to address a broader range of hallucinations. We aim to resolve image hallucinations more comprehensively and to develop quantitative evaluation metrics and benchmarks for this purpose.

\appendix

\bibliographystyle{named}

\begin{thebibliography}{}

\bibitem[\protect\citeauthoryear{Betker \bgroup \em et al.\egroup }{2023}]{betker2023improving}
James Betker, Gabriel Goh, Li~Jing, Tim Brooks, Jianfeng Wang, Linjie Li, Long Ouyang, Juntang Zhuang, Joyce Lee, Yufei Guo, et~al.
\newblock Improving image generation with better captions.
\newblock {\em Computer Science. https://cdn. openai. com/papers/dall-e-3. pdf}, 2(3):8, 2023.

\bibitem[\protect\citeauthoryear{Borgeaud \bgroup \em et al.\egroup }{2022}]{NLPimprove}
Sebastian Borgeaud, Arthur Mensch, Jordan Hoffmann, Trevor Cai, Eliza Rutherford, Katie Millican, George~Bm Van Den~Driessche, Jean-Baptiste Lespiau, Bogdan Damoc, Aidan Clark, et~al.
\newblock Improving language models by retrieving from trillions of tokens.
\newblock In {\em International conference on machine learning}, pages 2206--2240. PMLR, 2022.

\bibitem[\protect\citeauthoryear{Brooks \bgroup \em et al.\egroup }{2023}]{instructpix2pix}
Tim Brooks, Aleksander Holynski, and Alexei~A Efros.
\newblock Instructpix2pix: Learning to follow image editing instructions.
\newblock In {\em Proceedings of the IEEE/CVF Conference on Computer Vision and Pattern Recognition}, pages 18392--18402, 2023.

\bibitem[\protect\citeauthoryear{Chen \bgroup \em et al.\egroup }{2022a}]{murag}
Wenhu Chen, Hexiang Hu, Xi~Chen, Pat Verga, and William~W Cohen.
\newblock Murag: Multimodal retrieval-augmented generator for open question answering over images and text.
\newblock {\em arXiv preprint arXiv:2210.02928}, 2022.

\bibitem[\protect\citeauthoryear{Chen \bgroup \em et al.\egroup }{2022b}]{reimagen}
Wenhu Chen, Hexiang Hu, Chitwan Saharia, and William~W Cohen.
\newblock Re-imagen: Retrieval-augmented text-to-image generator.
\newblock {\em arXiv preprint arXiv:2209.14491}, 2022.

\bibitem[\protect\citeauthoryear{Hertz \bgroup \em et al.\egroup }{2022}]{hertz2022prompt}
Amir Hertz, Ron Mokady, Jay Tenenbaum, Kfir Aberman, Yael Pritch, and Daniel Cohen-Or.
\newblock Prompt-to-prompt image editing with cross attention control.
\newblock {\em arXiv preprint arXiv:2208.01626}, 2022.

\bibitem[\protect\citeauthoryear{Huang \bgroup \em et al.\egroup }{2023}]{survey}
Lei Huang, Weijiang Yu, Weitao Ma, Weihong Zhong, Zhangyin Feng, Haotian Wang, Qianglong Chen, Weihua Peng, Xiaocheng Feng, Bing Qin, et~al.
\newblock A survey on hallucination in large language models: Principles, taxonomy, challenges, and open questions.
\newblock {\em arXiv preprint arXiv:2311.05232}, 2023.

\bibitem[\protect\citeauthoryear{Lin \bgroup \em et al.\egroup }{2021}]{lin2021truthfulqa}
Stephanie Lin, Jacob Hilton, and Owain Evans.
\newblock Truthfulqa: Measuring how models mimic human falsehoods.
\newblock {\em arXiv preprint arXiv:2109.07958}, 2021.

\bibitem[\protect\citeauthoryear{Liu \bgroup \em et al.\egroup }{2024}]{regular}
Bingbin Liu, Jordan Ash, Surbhi Goel, Akshay Krishnamurthy, and Cyril Zhang.
\newblock Exposing attention glitches with flip-flop language modeling.
\newblock {\em Advances in Neural Information Processing Systems}, 36, 2024.

\bibitem[\protect\citeauthoryear{Robertson}{2024}]{verge24}
Adi Robertson.
\newblock Google's ai gemini generates inaccurate historical information.
\newblock \url{https://www.theverge.com/2024/2/21/24079371/google-ai-gemini-generative-inaccurate-historical}, February 2024.
\newblock Accessed: July 8, 2024.

\bibitem[\protect\citeauthoryear{Rombach \bgroup \em et al.\egroup }{2022}]{StableDiffusion}
Robin Rombach, Andreas Blattmann, Dominik Lorenz, Patrick Esser, and Bj{\"o}rn Ommer.
\newblock High-resolution image synthesis with latent diffusion models.
\newblock In {\em Proceedings of the IEEE/CVF conference on computer vision and pattern recognition}, pages 10684--10695, 2022.

\bibitem[\protect\citeauthoryear{Saharia \bgroup \em et al.\egroup }{2022}]{imagen}
Chitwan Saharia, William Chan, Saurabh Saxena, Lala Li, Jay Whang, Emily~L Denton, Kamyar Ghasemipour, Raphael Gontijo~Lopes, Burcu Karagol~Ayan, Tim Salimans, et~al.
\newblock Photorealistic text-to-image diffusion models with deep language understanding.
\newblock {\em Advances in neural information processing systems}, 35:36479--36494, 2022.

\bibitem[\protect\citeauthoryear{Wu \bgroup \em et al.\egroup }{2023}]{wu2023self}
Tsung-Han Wu, Long Lian, Joseph~E Gonzalez, Boyi Li, and Trevor Darrell.
\newblock Self-correcting llm-controlled diffusion models.
\newblock {\em arXiv preprint arXiv:2311.16090}, 2023.

\bibitem[\protect\citeauthoryear{Yang \bgroup \em et al.\egroup }{2023}]{yang2023paint}
Binxin Yang, Shuyang Gu, Bo~Zhang, Ting Zhang, Xuejin Chen, Xiaoyan Sun, Dong Chen, and Fang Wen.
\newblock Paint by example: Exemplar-based image editing with diffusion models.
\newblock In {\em Proceedings of the IEEE/CVF Conference on Computer Vision and Pattern Recognition}, pages 18381--18391, 2023.

\bibitem[\protect\citeauthoryear{Yasunaga \bgroup \em et al.\egroup }{2022}]{racm3}
Michihiro Yasunaga, Armen Aghajanyan, Weijia Shi, Rich James, Jure Leskovec, Percy Liang, Mike Lewis, Luke Zettlemoyer, and Wen-tau Yih.
\newblock Retrieval-augmented multimodal language modeling.
\newblock {\em arXiv preprint arXiv:2211.12561}, 2022.

\bibitem[\protect\citeauthoryear{Ye \bgroup \em et al.\egroup }{2023}]{ipadapter}
Hu~Ye, Jun Zhang, Sibo Liu, Xiao Han, and Wei Yang.
\newblock Ip-adapter: Text compatible image prompt adapter for text-to-image diffusion models.
\newblock {\em arXiv preprint arXiv:2308.06721}, 2023.

\bibitem[\protect\citeauthoryear{Yu \bgroup \em et al.\egroup }{2023}]{NLPimprove2}
Wenhao Yu, Zhihan Zhang, Zhenwen Liang, Meng Jiang, and Ashish Sabharwal.
\newblock Improving language models via plug-and-play retrieval feedback.
\newblock {\em arXiv preprint arXiv:2305.14002}, 2023.

\end{thebibliography}

\end{document}